\documentclass[letterpaper, 10 pt, conference]{ieeeconf}  

\IEEEoverridecommandlockouts                              

\usepackage{graphics} 
\usepackage{epsfig} 
\usepackage{mathptmx} 
\usepackage{times} 
\usepackage{amsmath} 
\usepackage{amssymb}  
\usepackage{pifont}
\usepackage{xcolor}
\usepackage{subcaption}
\usepackage{adjustbox}
\usepackage{array}
\usepackage{multirow, booktabs}
\usepackage{makecell}
\usepackage[mathscr]{euscript}
\usepackage{float}
\usepackage{gensymb}

\usepackage[bookmarks=true, hidelinks]{hyperref}    
\hypersetup{
     colorlinks   = true,
     urlcolor=black
}

\usepackage{algpseudocode}
\usepackage{cite}

\newcommand{\cmark}{\ding{51}}%
\newcommand{\xmark}{\ding{55}}%

\title{\LARGE \bf
Scene-level Tracking and Reconstruction without Object Priors
}

\author{Haonan Chang and Abdeslam Boularias$^{1}$
\thanks{$^{1}$Both authors are with the Department of Computer Science of
        Rutgers University, NJ, USA. This work is supported by NSF awards 1734492, 1846043, and 2132972.}%
}

\begin{document}

\maketitle
\thispagestyle{empty}
\pagestyle{empty}

\begin{abstract}
We present the first real-time system capable of tracking and reconstructing, individually, every visible object in a given scene, without any form of prior on the rigidness of the objects, texture existence, or object category. In contrast with previous methods such as Co-Fusion and MaskFusion that first segment the scene into individual objects and then process each object independently, the proposed method dynamically segments the non-rigid scene as part of the tracking and reconstruction process. When new measurements indicate topology change, reconstructed models are updated in real-time to reflect that change. Our proposed system can provide the live geometry and deformation of all visible objects in a novel scene in real-time, which makes it possible to be integrated seamlessly into numerous existing robotics applications that rely on object models for grasping and manipulation. The capabilities of the proposed system are demonstrated in challenging scenes that contain multiple rigid and non-rigid objects. 
Supplementary material, including video, can be found at  
 \href{https://github.com/changhaonan/STAR-no-prior}{\texttt{\textcolor{blue}{https://github.com/changhaonan/STAR-no-prior}}}. 

\end{abstract}

\section{INTRODUCTION}
Robots are increasingly deployed in unstructured environments such as households, warehouses, and workshops. The large variety of object types, shapes, and textures that are encountered in such environments makes it virtually impossible for robots to always rely on prior models of the objects for grasping, dexterous manipulation, and motion planning.
Moreover, visual demonstrations for teaching robots new tasks in real-world setups are often performed in complex scenes that contain a number of unknown and novel objects. The objects are often non-rigid and partially occluded during the demonstrations. Consequently, tracking and 3D reconstruction of previously unseen objects is an important component of any intelligent robotic system. 

Thanks to the recent availability of low-price depth-sensing cameras, tracking and 3D reconstruction of objects became two popular topics in robot vision~\cite{sui2020geofusion}. 

These two problems are however interrelated; reconstructed 3D object models facilitate subsequent tracking of the object, while object tracking is often essential for the reconstruction of its 3D model, which is highly challenging in the absence of object priors, as it is often the case. 
Therefore, these two problems need to be solved jointly in a unified framework.

The Scene-level Tracking and Reconstruction (\textbf{STAR}) problem brings the above challenge to a scene scale: Given a sequence of RGB-D images of an entire scene, can we reconstruct the geometry model and track the pose and deformation of each moving object in the scene simultaneously?

One approach to solving this problem is to first segment the scene images into point clouds of individual objects and then treat them separately during tracking and reconstruction~\cite{Runz2017, Runz2019, xu2019mid}. This approach however heavily relies on the existence of a pre-trained neural network (such as {\it MaskRCNN}) as a prior for segmentation, which limits its applicability in settings with entirely new types of objects.

Another approach is to reverse the order of segmentation and tracking, reconstruction.
First, reconstruct the non-rigid scene geometry in its entirety, using all the RGB-D images of the scene. Then, segment the reconstructed 3D scene model into individual objects. A major challenge in the first step lies in handling topological changes that affect the feasibility of the second step. Topological changes occur for example when separate objects, or parts of an object, become physically connected and vice versa.

Topological changes in tracking and reconstruction of non-rigid objects are handled differently depending on the type of the geometry model that is adopted by the method, such as {\it TSDF} (Truncated Signed Distance Field) and {\it Surfels} (surface elements).
For TSDF-based methods, some methods reset the entire model whenever a major topological change is detected~\cite{Dou2016,Dou2017}. Other methods use the status of spatial compression to determine areas where topology changes and replace those areas~\cite{Zampogiannis2018, Li2020a}. These methods however either rely on complex and ultra-precise sensors~\cite{Dou2016,Dou2017}, or depend on accurate texture-based registration \cite{Zampogiannis2018,Li2020a}, which makes them difficult to use in general robotics applications. 
A recent work based on a {\it killing-field} approximation reduces the dependency on texture while handling large topological change, but its use of an SDF-based geometry makes object segmentation very difficult and non-trivial~\cite{Slavcheva2020,slavcheva2018sobolevfusion}. 
Surfel-based models are more flexible for handling topology changes in non-rigid reconstruction~\cite{Keller2013,Gao2019}. Instead of resetting the entire model whenever a major topology change is detected and thus losing important information, surfel-based methods can locally re-initialize the geometry around affected areas. This can be achieved because, unlike TSDF models, surfel models do not have any internal constraints, which makes the removal and appending of point clouds relatively easy. Furthermore, a surfel representation can be beneficial in geometric separation leading to object segmentation. 

In this paper, we propose a novel system that can solve the scene-level tracking and reconstruction problem without any object or category prior, or any assumption regarding the rigidness or texture of the objects. The contributions of this work can be summarized as follows:
\noindent \textbf{(1) A new surfel-based method for non-rigid scene reconstruction that can handle topology change.} Our proposed non-rigid geometry reconstruction pipeline improves the multi-view SurfelWarp technique~\cite{Gao2019} by introducing a local re-initialization strategy, which is used in our approach to detect and handle topology changes. Compared to existing solutions to this problem, ours does not rely on texture-based registration or a category-dependent refiner.  A comparison between our proposed pipeline and multi-view SurfelWarp~\cite{Gao2019} clearly illustrates the advantages of our approach.
\noindent \textbf{(2) The first real-time scene-level, tracking and reconstruction solution that does not use any object prior.} Existing solutions such as MaskFusion require objects to be rigid and rely on pre-trained neural networks as priors for segmenting scenes into objects. To our best knowledge, our system is the first that solves this problem and returns an individual model for each moving object in the scene without any prior information about the objects, their rigidness, or texture. 
The proposed system is tested over a set of RGB-D images of challenging scenes, with various types of objects. 

\begin{table}[ht!]
\centering
\caption{\small Properties of different real-time, dense-SLAM  scene reconstruction systems.}
\begin{adjustbox}{width=0.95\linewidth}
\begin{tabular}{ccccccc}
\toprule
\multirow{2}{*}{Method} & Category & Dynamic & Non-rigid & Topology & Segmentation & Texture\\
& free & scene & objects & change &  & free\\ 
\cmidrule{1-7}
SLAM++  & \color{red}\xmark  & \color{red}\xmark & \color{red}\xmark   &\color{red}\xmark   &\color{red}\xmark   &\color{red}\xmark   \\ 
\specialrule{0em}{0.5pt}{0.5pt}
DynamicFusion \cite{Newcombe}  & \color{green}\cmark &  \color{green}\cmark  & \color{green}\cmark  &  \color{red}\xmark &  \color{red}\xmark  &  \color{green}\cmark   \\ 
\specialrule{0em}{0.5pt}{0.5pt}
Volume Deform  & \color{green}\cmark &  \color{green}\cmark  & \color{green}\cmark  &  \color{red}\xmark &  \color{red}\xmark  &  \color{red}\xmark   \\
\specialrule{0em}{0.5pt}{0.5pt}
SurfelWarp \cite{Gao2019}& \color{green}\cmark&  \color{green}\cmark& \color{green}\cmark&  \color{red}\xmark& \color{red}\xmark&  \color{green}\cmark\\ 
\specialrule{0em}{0.5pt}{0.5pt}
Fusion4D \cite{Dou2016}& \color{green}\cmark&  \color{green}\cmark& \color{green}\cmark&  \color{green}\cmark& \color{red}\xmark&  \color{red}\xmark\\ 
\specialrule{0em}{0.5pt}{0.5pt}
Motion2Fusion \cite{Dou2017} & \color{green}\cmark&  \color{green}\cmark& \color{green}\cmark&  \color{green}\cmark& \color{red}\xmark&  \color{red}\xmark \\ 
\specialrule{0em}{0.5pt}{0.5pt}
Functon4D \cite{Yu}& \color{red}\xmark&  \color{green}\cmark& \color{green}\cmark&  \color{green}\cmark& \color{red}\xmark&  \color{green}\cmark \\ 
\specialrule{0em}{0.5pt}{0.5pt}
TCAFusion \cite{Li2020a} &  \color{green}\cmark&  \color{green}\cmark& \color{green}\cmark&  \color{green}\cmark& \color{red}\xmark&  \color{red}\xmark\\ 
\specialrule{0em}{0.5pt}{0.5pt}
Co-fusion \cite{Runz2017} & \color{red}\xmark & \color{green}\cmark& \color{red}\xmark& \color{green}\cmark& \color{green}\cmark& \color{green}\cmark \\ 
\specialrule{0em}{0.5pt}{0.5pt}
MaskFusion \cite{Runz2019}& \color{red}\xmark & \color{green}\cmark& \color{red}\xmark& \color{green}\cmark& \color{green}\cmark& \color{green}\cmark\\ 
\specialrule{0em}{0.5pt}{0.5pt}
MidFusion \cite{xu2019mid}  & \color{red}\xmark & \color{green}\cmark& \color{red}\xmark& \color{green}\cmark& \color{green}\cmark& \color{green}\cmark\\ 
\specialrule{0em}{0.5pt}{0.5pt}
\textbf{Ours}  & \color{green}\cmark & \color{green}\cmark & \color{green}\cmark  &\color{green}\cmark & \color{green}\cmark  &   \color{green}\cmark \\ 
\bottomrule
\end{tabular}
\end{adjustbox}
\label{taxonomy}
\end{table}

\section{RELATED WORKS}
We discuss in the following some of the recent related techniques. Table~\ref{taxonomy} shows a taxonomy of these techniques.

{\bf Simultaneous tracking and reconstruction.}
Simultaneous tracking and reconstruction is
an important problem in robotic manipulation, since manipulation planning and learning algorithms often require geometric models of the objects and their poses~\cite{Runz2017}. Unlike earlier works centered on single object tracking and reconstruction, 
more recent techniques focus on dealing with multiple objects simultaneously. Examples of such techniques include Co-fusion~\cite{Runz2017}, MaskFusion~\cite{Runz2019} and MidFusion~\cite{xu2019mid}. These techniques require priors for segmenting the given scene initially into multiple objects and then tracking and reconstructing each object separately. Object priors are not always available, and initial segmentation errors can lead to significant tracking and reconstruction errors later. Moreover, these techniques assume that the objects are rigid.

{\bf Dynamic Scene Reconstruction.}
Dynamic scene reconstruction typically uses a static geometric model and a deformation field to describe a deformable object or a dynamic scene. Existing techniques require solving joint optimization problems, which are computationally expensive~\cite{Collet,dou20153d}. DynamicFusion~\cite{Newcombe} is a parallel GPU-based solution that solves this problem more efficiently and that can be considered as the first online non-rigid reconstruction system. However, DynamicFusion cannot handle topological changes in measurements. More recent techniques such as Fusion4D~\cite{Dou2016}, Motion2Fusion~\cite{Dou2017}, and others~\cite{Li2020a,Zampogiannis2018}, try to solve this problem with texture-based, learning-based registration or global re-initialization. Some of these works achieved excellent reconstruction results. But to the best of our knowledge, all existing topology-aware non-rigid reconstruction methods either rely on texture-rich measurements or topology-prior for registration, or on category-level priors (such as a human or a pre-trained network) for model refinement.
In contrast with existing solutions, we exploit an intriguing property of surfels that makes them easy to remove or append locally, to perform local re-initialization. Our proposed method makes no assumption on the existence of texture or category-level prior to the target scene.

{\bf Surfel-based Reconstruction.}
Surfels (surface elements) are first proposed by Pfister et al.~\cite{pfister2000surfels} as rendering primitives. A surfel is a zero-dimensional n-tuple that can approximate a local surface. Keller et al.~\cite{Keller2013} defined a surfel as a tuple of a 3D position, a normal, and a radius and first used surfels for real-time reconstruction. Then SurfelWarp~\cite{Gao2019} was proposed to extend the use of surfels to non-rigid 
scene reconstruction. Using surfels for geometry representation in non-rigid reconstruction has the advantage of faster processing and smaller memory usage~\cite{Gao2019}. However, current surfel-based methods do not handle large topological change (e.g., surface splitting) without global re-initialization. We found in this work that
surfels are efficient for local re-initialization because they can be structured as a simple unordered list of points. Compared to TSDF-based models, removing failure parts and appending new geometry locally is much easier for surfel-based models. Meanwhile, this property also makes it easy to split a model into models for different objects, which is another key requirement for scene-level segmentation. 
\section{problem formulation and background} \label{problem}
We consider the problem of simultaneous tracking and reconstruction of all the objects that are present in a given dynamic scene, using as inputs a sequence of RGB-D images taken from $K$ different camera poses. The objects and their number are completely unknown {\it a priori}. The objects can also be non-rigid. A depth map in a given RGB-D image is denoted as $d:\Omega \rightarrow \mathbb{R}$ where $\Omega$ is the set of pixels in the image, and $d(u)$ is the depth of pixel $u = (x,y) \in \Omega$. Similarly, we denote the RGB image as $c:\Omega \rightarrow [0,1]^3$.

The output of the proposed system at each time-step $t$ is a set containing a {\it Surfel-based} geometry $S_i^t$ for each individual object $i$ in the scene, and a corresponding deformation graph $G_i^t$. Surfel-based geometry $S_i^t$ is a set of surfels $s_j$. A surfel $s_j$ is defined as $s_j = (v_j, n_j, c_j, r_i)$, where $v_j, n_j, c_j, r_j$  are respectively the 3D coordinates, normal, color and radius of surfel  $s_j \in S_i^t$. Different from previous methods \cite{Newcombe},\cite{Gao2019}, which keep a canonical geometry model and a live geometry model, we only keep the latest geometry model, because we only care about the current geometry of the scene and objects.
Deformation graph $G_i^t$ is defined by a set of nodes $\{g_i^t\}$ that correspond to 3D points belonging to the same topology (i.e., same object). Each node in the graph is connected to its nearest-neighbors. Nodes and edges in $G_i^t$ are added or dropped dynamically as topology changes are detected. Deformation graph $G_i^t$ of object $i$ is accompanied with a {\it warp field} $W_i^t$. A warp field is defined as $W = \{ [p_j \in \mathbb{R}^3, \delta_j \in \mathbb{R}^+, T_j \in SE(3)]\}$, wherein $j$ is the node index in the accompanying graph, $p_j$ is the 3D point that corresponds to node $j$, $\delta_j$ is a the node's radius of influence, and $T_j$ is the 6-D transformation defined on node $g_j$. Here $T_j$ is represented by a {\it dual quaternion} $q_j$ for smooth interpolation~\cite{Kavan2007}. Warp field $W$ is used to describe the deformation between two consecutive time steps. For each surfel $s = (v, n, c, r) \in S$, we compute its 6-D transformation $\bar W(s)$ based on warp field $W$ by applying the formula
\begin{equation}
    \bar W(s) = normalize(\sum_{k\in N(s)} \omega_k(v,p_k)q_k)
    \label{warp_field}
\end{equation}
Here, $N(s)$ denotes the set of nodes that are the neighbors of the surfel $s$ (see Sec.~\ref{representation} and Fig.~\ref{graph_illustration} for details). $\omega(s)$ is an interpolation parameter, defined as $\omega(s) = \exp\big(\left\Vert  v - p_k \right\Vert^2_2 / (2\delta_k^2)\big)$, $v$ is 3-D position of surfel $s$.
The local transformation $\bar W(s)$ is then used to describe the deformation of surfel $s$ as follows,
\begin{equation}
  \begin{split}\label{warp}
    \bar W^t(s) v &=  v_{warp},\\ rotation\big(\bar W(s)\big) n &=  n_{warp},
  \end{split}
\end{equation}

wherein $v,n$ are vertex and normal of $s$ before deformation, and $v_{warp},n_{warp}$ are those after warping.

\section{PROPOSED APPROACH}
\subsection{Overview}
\begin{figure}[ht]
 \center
  \includegraphics[width=0.48\textwidth]{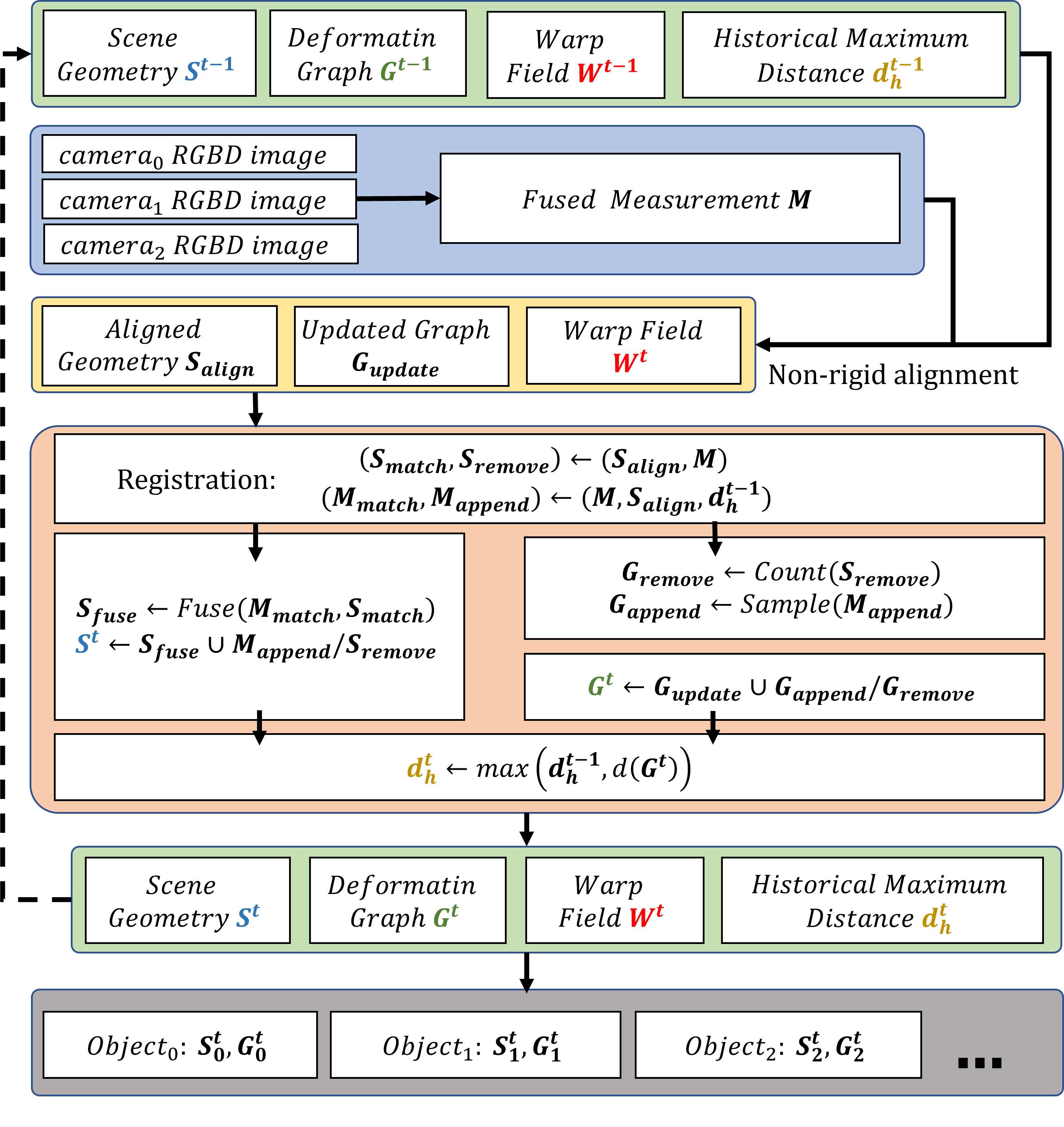}
  \caption{\small Overview of the proposed pipeline}
  \label{overview_pipeline}
\end{figure}

An overview of the proposed system is shown in Fig.\ref{overview_pipeline}. This pipeline is divided into four main components: (1) measurement fusion (shown in blue), (2) non-rigid alignment (yellow), (3) geometry and deformation graph update (pink), and (4) topology separation (grey). The green box refers to the entire scene representation model $(S^{t-1},G^{t-1},W^{t-1},d^{t-1}_h)$ at time-step $t-1$ as input to the system, and as its output at time-step $t$. At each time-step, RGB-D measurements of the scene from different cameras are fused into a single surfel-based geometry $M$, defined as a set of surfels (Sec.~\ref{problem}). Current model $(S^{t-1},G^{t-1},W^{t-1},d^{t-1}_h)$ is aligned with in-coming fused measurement $M$ through non-rigid alignment (Sec.~\ref{optimization}), which results in a new scene geometry $S_{align}$. Then, a registration between $S_{align}$ and $M$ is performed (Sec.~\ref{fusion}). The parts of $S_{match}$ and $M_{match}$ that match together are fused. The unmatched parts of the geometry from the existing model (i.e., outliers) are removed, and the unregistered measurement (newly observed parts of the scene) is appended to the model. In parallel to the updates of the geometry model, deformation graph $G^{t-1}$ is updated by removing nodes that are out of track and expanding the graph with nodes corresponding to newly observed points. Finally, historical maximum distance $d^{t-1}_h$ is updated. In addition to the updated scene model, the system returns a set of local models $(S_i^t, G_i^t)$, one for each detected object after applying a topology-based segmentation.

\subsection{Deformation model representation} \label{representation}
The proposed method uses a surfel-based model instead of the TSDF model which is the main-stream model used for non-rigid scene reconstruction. We found from our investigation that a surfel-based representation is better than TSDF when facing frequent topology changes. TSDF is efficient at maintaining local topology because each surface is determined by all its nearby voxels, which also implies however that local topology is hard to change. When a local geometry is lost, due to occlusion, the entire TSDF model needs to be reset to fix the local problem. In contrast, a surfel-based representation has no internal constraints over topology. Therefore, failures can be fixed locally. 

\begin{figure}[ht]
 \center
  \includegraphics[width=0.48\textwidth]{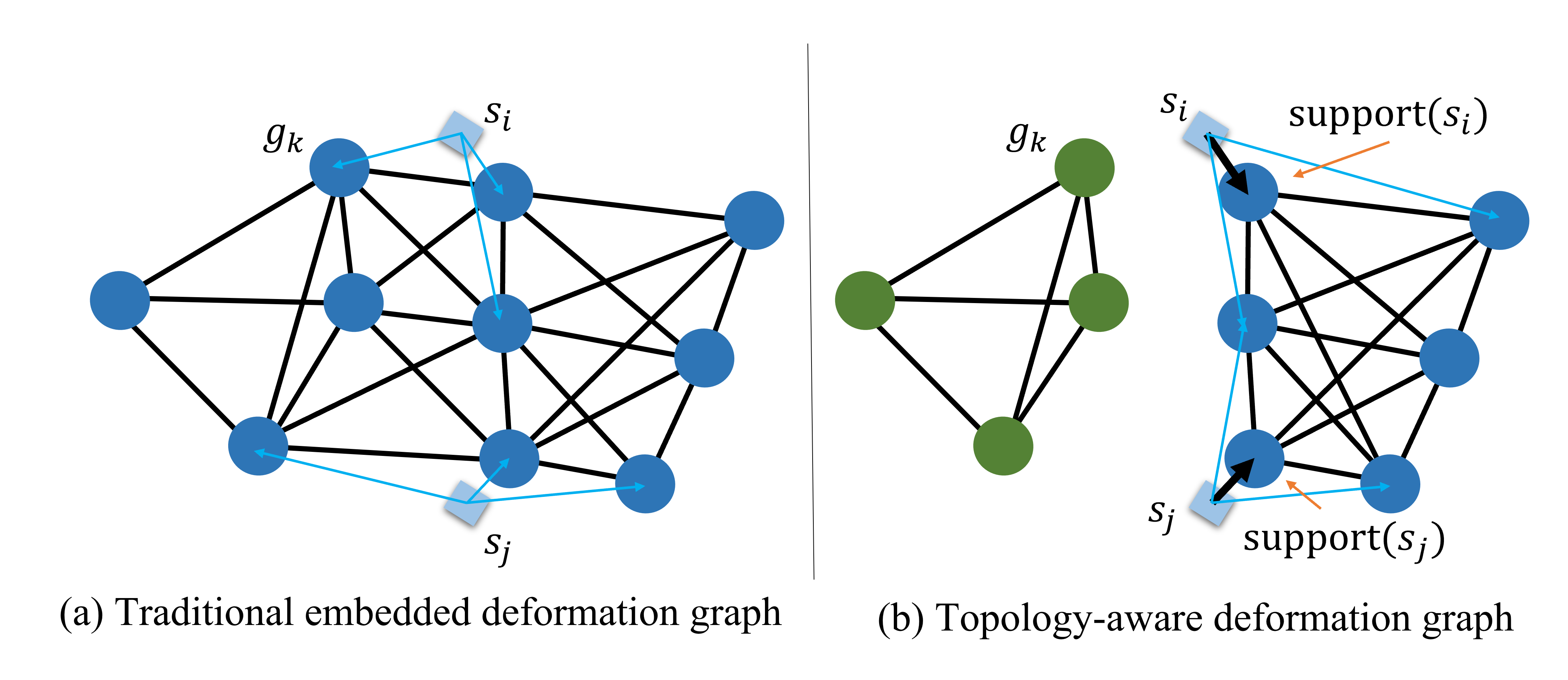}
  \caption{\small Proposed topology-aware deformation graph. In traditional embedded deformation graphs, neighbors of a node (such as $g_k$) are decided by their current Euclidean distances, and neighbors of a surfel (such as $s_i$ and $s_j$) are also defined by their Euclidean distance to the nodes in the graph. In our proposed deformation graph, neighbors of a node are determined by a historical maximum distance $d_t$. Also, neighbors of a surfel can only be selected from the same sub-graph as its support node ($\textrm{support}(s_i)$ and $\textrm{support}(s_j)$).}
  \label{graph_illustration}
\end{figure}

On the other hand, surfel representations have the 
disadvantage of not carrying any topology information.
A typical surfel-based model cannot distinguish between two surfels belonging to different objects. 
Thus, we propose a hybrid model that combines a surfel-based geometry $S$ with a topology-aware deformation graph $G$. 
As shown in Fig.~\ref{graph_illustration}, for each surfel $s_i \in S$, we assign a node $g_j = \textrm{support} (s_i,G)$ from $G$ as its support node. Support node of $s_i$ is defined as the nearest point in $G$ to $s_i$ when it joined the geometry $S$, using the Euclidean distance. Each surfel is assigned to a unique support node while one support node typically supports numerous surfels.
Graph $G$ is dynamically decomposed into several topologies (i.e., connected components), one per object. Surfel $s_i$ is assigned to the same topology as its supporting node $g_j$. When $g_j$ is removed, all surfels supported by $g_j$ are also removed from $S$. Deformation graph $G$ is the skeleton of the 3D model, and surfel set $S$ is its skin. The topology of the surfel set $S$ is represented by its skeleton $G$.

Deformation graph $G$ is dynamic, its nodes and edges change over time. Each node in $G$ is connected to its $k$-nearest neighboring nodes in $G$, using the \textbf{historical maximum distance} $d_h$ as a metric, which we define here as:

\begin{equation}
    \forall g_i, g_j \in G, d_h(g_i, g_j) = max_{t\in T}\{ d^t(g_i, g_j)\},
\end{equation}
wherein $T$ is the number of time-steps (or frames) in the sequence of images, and $d^t(g_i, g_j)$ is the Euclidean distance between nodes $g_i$ and $g_j$ at that time step $t$. The historical maximum distance proposed here is better than the Euclidean distance in terms of determining topology. 
Two nodes $g_i$ and $g_j$ belonging to different objects can be close to each other for arbitrarily long periods, which happens for example when one of the objects is resting on the other. But if the two nodes have been observed to be far away from each other at any moment in the past frames, then they should not be neighbors in $G$. 

Given deformation graph $G$, we define the set $N(s)$ of neighbors of surfel $s\in S$ as the $k'$-nearest neighbors of $s$ among the $k$-nearest neighbors of $g$, the support node of $s$. In other terms, 
$N(s) = k'\textrm{NN}\big(s,k\textrm{NN}(g,G)\big)$, with $g = \textrm{support}(s,G)$ and $k'\leq k$. Unlike the $k$-nearest neighbors of $g$ selected using the historical maximum distance, the $k'$-nearest neighbors of $s$ are obtained using the Euclidean distance at the latest time step. Because all neighbors of $s$ are selected among the neighbors of its support node in $G$, they necessarily belong to the same connected component in $G$, i.e., same topology.

\noindent In summary, we use $S$ (surfel-based geometry), $G$ (deformation graph), $W$ (warp field), $d_h$ (historical maximum distance) to describe the deformable model of the entire scene.

\subsection{Initialization} \label{initialization}
$S,G,W,d_h$ are initialized at the first frame. The geometry $S$ is initialized with the measurement $M$ from the first frame. The nodes of deformation graph $G$ are uniformly sampled from those initializing surfels spatially with the same algorithm as DynamicFusion \cite{Newcombe}. The initializing neighbor set $N(g)$ of every node $g$ in $G$ is their kNN set by Euclidean distance at the first frame. 
The historical maximum distance $d_h$ is initialized by the Euclidean distance at the starting frame. The warp-field $W$ associated with the deformation graph $G$ is initialized to an identity transformation.

\subsection{Measurement}
The measurement acquisition system in our pipeline is similar to that of  Function4D~\cite{Yu}, a sparse multi-camera system consisting of three RealSense-415 RGB-D cameras. Multi-view measurement fusion is achieved through a TSDF-based fusion pipeline similar to the one used in~\cite{Dou2016,Dou2017}. It is worth noting that we use TSDF here instead of the surfel-based representation because surfel models are more sensitive to distortion and calibration error across different cameras, and correcting for these errors is hard and computationally expensive. 
This problem with surfels was addressed in~\cite{Collet} by projecting the point cloud to local planes, but this solution is still computationally inefficient and relies on a multi-camera system with $106$ cameras. Alternatively, a TSDF model can dramatically decrease measurement noise and achieve a good smoothness between measurements from different sources with a low computational cost. 

Since we rely on the measurement to detect areas where topology changes (Sec. \ref{fusion}), we only keep the best measurement. According to \cite{Whelan2015}, the noise level of depth measurement at a point is proportional to the angle between the local normal and the camera view direction. Thus, we discard measurements that have a viewing angle larger than $70 \degree$.

After each measurement fusion, we use the Raycast algorithm described in KinectFusion~\cite{izadi2011kinectfusion} to transform the TSDF into a measurement surfel model $M$ for further processing.

\subsection{Non-rigid alignment} \label{optimization}
We show here how to compute the non-rigid warp field by solving a massive optimization problem on a GPU. For this step, we principally follow the method of DynamicFusion~\cite{Newcombe} and SurfelWarp~\cite{Gao2019}, and adapt them to our surfel-based model and multi-view setting.
The key idea of non-rigid warp field estimation is to align the geometry model $S^{t-1}$ to the current measurement $M$. This can be formulated as the following optimization problem,
\begin{equation}
\min_{W} E_{total}(W) \textrm{ with } E_{total}(W) = E_{depth} (W) + \lambda E_{reg} (W).  \label{energy}
\end{equation}
Here, $E_{depth}$ is used to align measurement $M$ and geometry $S$. $E_{reg}$ is used to regulate the deformation within each topology favoring rigidity. $\lambda$ is a balancing parameter. 
Parameter $\lambda$ is typically small because strong regularization constraints prevent topology from separation.
Furthermore, $E_{depth}$ is defined as follows,
\begin{equation}
    E_{depth}(W) = \sum_{i=1}^{K} \sum_{(s^{t-1},s_{M}) \in P_i} \big(n_{M}^T (v^{t-1} - v_{M})\big)^2 .
    \label{depth_energy}
\end{equation}
Here, $K$ is the number of cameras, $P_i$ is a set of pairs of measured depths $s_{M}$ from each camera pose $i$ and their corresponding rendered models $S^{t-1}$. The normals of $s_M$ and $s^{t-1}$ are denoted by $n_{M}$ and $n^{t-1}$, while $v^{t-1}$ and $v_{M}$ are the vectors containing the 3D vertices of $s^{t-1}$ and $s_{M}$.

Rendered models are obtained by rendering the global geometry $S^{t-1}$ under different camera views. 
Each rendered image has the same size as its corresponding measurement depth image. 
Thus, for each pixel $u=(u_x,u_y)$ on the rendered geometry map, we search within a small neighborhood $(u_x \pm \sigma, u_y \pm \sigma)$ for the best correspondence in terms of position and normal difference. If their position distance or normal difference are above thresholds $\gamma_{distance}$ and $\gamma_{normal}$, this pair will be discarded from the cost computation in Equation~\ref{depth_energy}.

As for the regulation term $E_{reg}$, it is defined as follows,
\begin{equation}
    E_{reg}(W) = \sum_{g_j \in G} \sum_{g_i \in N(g_j)} \left\Vert  T_j p_j - T_i p_i \right\Vert^2_2 . 
\end{equation}
This regulation term is used to constrain deformations within each topology to be as rigid as possible. The topology structure is described by $N(g_j)$, the set of neighbors of node $g_j$ in the deformation graph. Neighbors are not defined as the nearest ones by Euclidean distance, but by a historical maximum distance $d_h$ (See Sec. \ref{representation}).

This optimization problem is a nonlinear least-squares problem. Thus, we solve it with the Gaussian-Newton algorithm. Every step  here is implemented with CUDA on single GPU efficiently, and the solution process runs on real-time.

The output of this step is the warp-field estimate in time step $t$, $W^t$. The aligned geometry $S_{align} = W^t S^{t-1}$. This deformation is defined in Equations~\ref{warp_field} and \ref{warp}. We also update the node position of the deformation graph $G_{update}$ with the following formulas:
\begin{equation}
\label{warp_graph}
  \begin{split} 
  g_{update} &\leftarrow \bar{W^t}(g^{t-1}) g^{t-1}\\
  \bar{W^t}(g^{t-1}) &= normalize(\sum_{k\in N(g^{t-1})} \omega_k(g^{t-1},p_k)q_k)
  \end{split}
\end{equation}

Here $\omega_k$ has the same definition as Equation~\ref{warp_field}. $N(g^{t-1})$ is the neighbor set of $g^{t-1}$ in graph $G^{t-1}$.

\subsection{Geometry and deformation graph update} \label{fusion}
\begin{figure}[ht]
 \center
  \includegraphics[width=0.48\textwidth]{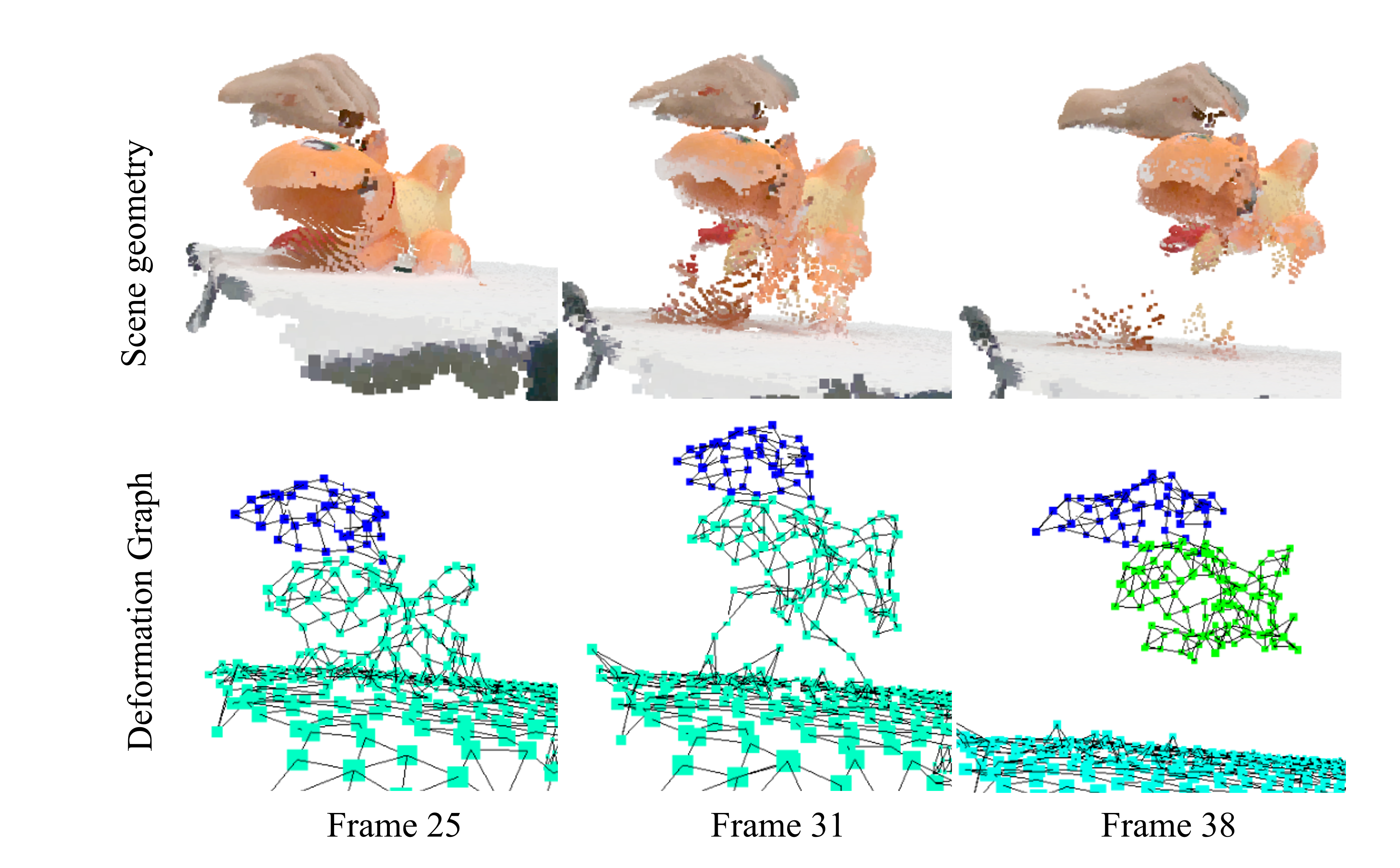}
  \caption{\small Change of the deformation graph when topological change is detected. At the beginning, the non-rigid plush toy belongs to the same topology as the table. The toy is then grasped and lifted. After the non-rigid alignment step, the upper part geometry of the toy is aligned to the current measurement. However, the remaining part of the toy still belongs to the same topology as the table, staying still at its original location. Since this part is out of track, its surfels are removed from the geometry. When enough surfels are removed due to their inconsistency with the free space, the removal of their support nodes is also triggered. After enough nodes in-between the deformation graphs of the toy and the table are deleted, the table and the toy's topologies are separated.
  }
  \label{topology_change}
  \vspace{-0.75cm}
\end{figure}

\begin{figure*}[ht]
    \begin{subfigure}[b]{\textwidth}
    \centering
    \includegraphics[width=\textwidth]{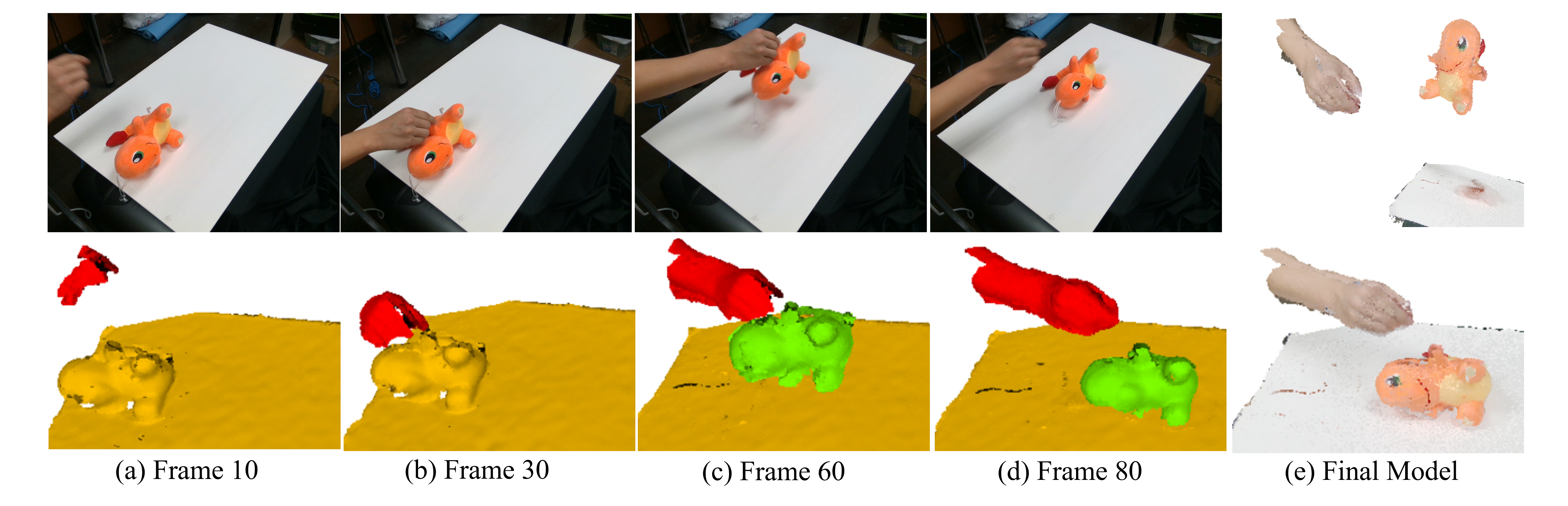}
    \vspace{-0.2cm}
    \caption{\footnotesize We lift a fire dragon plush toy from the table and release it. The first row is the RGB image from camera 0, the second row is the corresponding segmentation result at that frame. The top right shows the separated models for each object. The bottom right is the scene geometry at the final frame.}
    \label{sequence_demonstrate}
    \end{subfigure}
    \begin{subfigure}[b]{\textwidth}
    \centering
    \includegraphics[width=\textwidth]{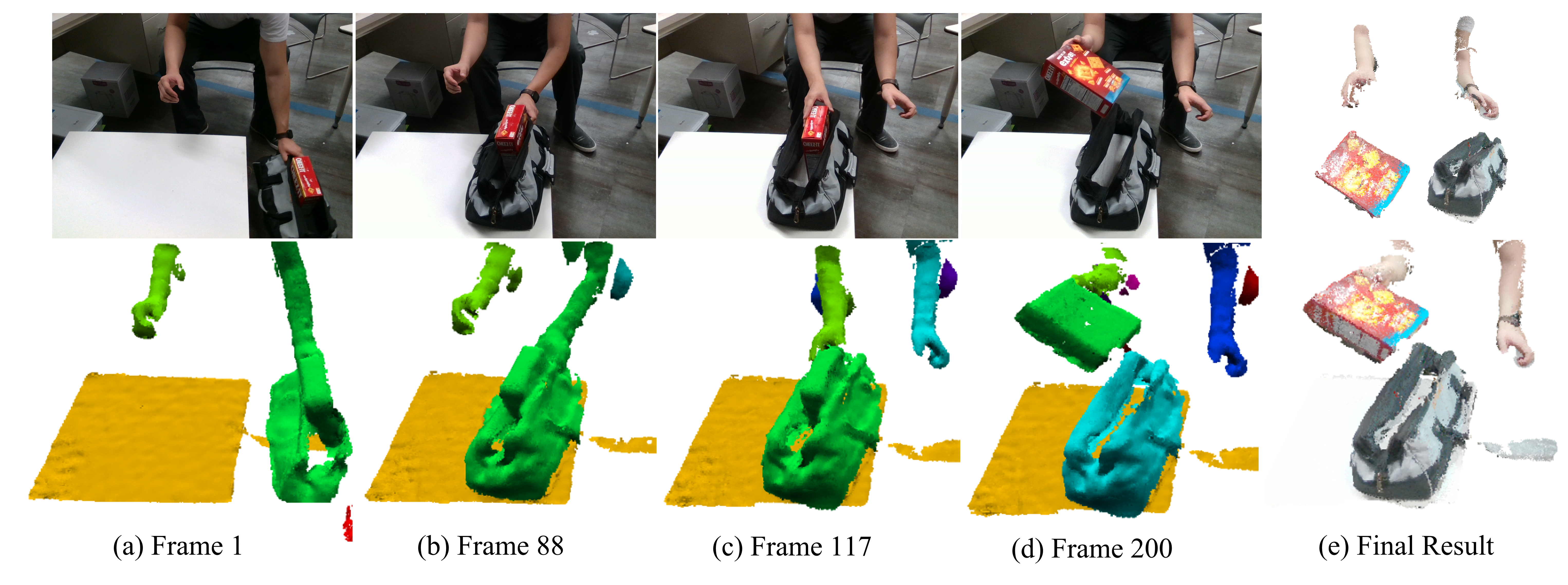}
    \vspace{-0.3cm}
    \caption{\footnotesize We use one hand to lift a non-rigid bag and put it on the table. Then, we pick a cracker box up from the bag with another hand.}
    \vspace{-0.1cm}
    \label{sequence_demonstrate_2}
    \end{subfigure}
    \caption{\small Experiments on two scenarios with a multitude of topological changes. Whenever a topology is discovered to be separate from others, it will be identified as an individual object. No object prior is used during the entire process. While such topological changes can relatively easily be detected by other techniques that make rigid-body assumptions, the non-rigidity of objects (i.e. the plush toy, hand, bag) makes the simultaneous reconstruction, tracking, and topology change handling extremely challenging.}
\vspace{-0.5cm}
\end{figure*}

There are four steps in this section: registration, fusion, appending, and removal.

In a nutshell, after the non-rigid alignment (See Sec. \ref{optimization}), the geometry from the last time step $S^{t-1}$ is warped to $S_{align}$, close to the current measurement $M$. However, there still exists a discrepancy between the aligned geometry $S_{align}$ and the measurement surfel $M$. There could be multiple reasons for this: measurement noises, newly observed surfaces, topology changes, etc.  Thus, we perform registration between $S_{align}$ and $M$. A match between $s_{align}$ and $s_M$ means the geometry is being observed in the current measurement. We need to fuse $s_M$ into $s_{align}$ to average the measurement noise. Meanwhile, if a given $s_M$ has no match in $S_{align}$, this $s_M$ can be a newly observed surface or a noise. If we verify this $s_M$ as newly observed geometry, we should append it to $S^t$. And when a given $s_{align}$ finds no correspondence to $M$, it is likely to be an outlier or just currently invisible. If it is an outlier,
then it should be removed from geometry $S^t$.

\textbf{Registration.}
The geometry registration pipeline is similar to SurfelWarp~\cite{Gao2019}, but in a multi-view version. We first render the aligned geometry $S_{align}$ and the fused measurement $M$ to each camera pose $k$ to generate an index map $\mathscr{I}_k$ and a depth map $D_k$. Each surfel will be projected to the camera's coordinates with corresponding camera intrinsic $I_k$ and camera pose $T_{camera}^k$. To avoid nearby surfels being projected to the same pixel in the index map. The index map is super-sampled by a scale of $4\times4$.
Then by comparing the index map $\mathscr{I}_k$ with depth measurement $D_k$, we can get the correspondence between geometry surfel $S_{align}$ and measurement surfel $M$.
Given a pixel position $u$ on the depth measurement, we search a $4\times 4$ area on the corresponding part of the index map. Among these points, we select the correspondence with the following criteria:
(1) Ignore surfels whose distance to the depth vertex is larger than $\gamma_{distance}$.
(2) Ignore surfels whose normal is far away from depth normal. i.e. $dot(n_{surfel}, n_{depth}) < \gamma_{normal}$.
(3) If there are multiple such surfels, choose the one that is nearest to the depth vertex.

\textbf{Fusion.} If such a correspondence between $s_{align}$ and $s_{M}$ is found, we will use the following formula to fuse them:
    $v_{fuse} \leftarrow  \frac{c_{align} v_{align} + c_{M} v_{M} }{(c_{align} + c_{M})},
    n_{fuse} \leftarrow  \frac{c_{align} n_{align} + c_{M} n_{M} }{(c_{align} + c_{M})}, r_{fuse} \leftarrow  \frac{c_{align} r_{align} + c_{M} r_{M} }{(c_{align} + c_{M})}, t_{fuse} \leftarrow t_{M}, c_{fuse} \leftarrow c_{align} + c_{M}$.
Here, $v,n,r,t,c$ are vertex, normal, radius, time-stamp and confidence of a surfel. The update is done in the above order sequentially.

\begin{align*}
    v_{fuse} & \leftarrow  \frac{c_{align} v_{align} + c_{M} v_{M} }{(c_{align} + c_{M})} \\
    n_{fuse} & \leftarrow  \frac{c_{align} n_{align} + c_{M} n_{M} }{(c_{align} + c_{M})} \\
    r_{fuse} & \leftarrow  \frac{c_{align} r_{align} + c_{M} r_{M} }{(c_{align} + c_{M})} \\
    t_{fuse} & \leftarrow t_{M}, c_{fuse} \leftarrow c_{align} + c_{M}
\end{align*}

\textbf{Append.} If no correspondence is found for a measurement surfel $s_{M}$, we will mark it as an appending candidate. Candidates under the following three cases will not be appended:

If $s_M$ is far away from existing geometry, it is likely to be a measurement noise. Furthermore, the multi-view camera setting brings a problem of different distortion models. In our practice, we find that there exists a shift in depth among different cameras even towards the same area. In case we create multiple surfaces for the same geometry, candidates too close to existing surfaces are thrown away. Finally, candidates supported by a \textit{compressed} node should not be appended. A node $g_i \in |G|$ can be defined as compressed if the following holds,
\begin{equation}
    \exists g_j \in |G|, d_h^t(g_i, g_j) > \gamma_{upper}, d^t(g_i, g_j) < \gamma_{lower},
\end{equation}
$d_h^t$ is the historical maximum distance, and $d^t$ is the Euclidean distance in the current frame.
A node becomes compressed when there exist some other nodes that are far away in history but currently near each other. This often happens when separated topologies approach each other (i.e., a hand approaching the table). To avoid having different topologies wrongly connected together by appending nodes, the appending process is not performed near these areas.  

In summary, the following criteria are imposed:
(1) Discard surfels $s$ whose distance to nearest deformation node is larger than $\gamma_{nn}$.
(2) Discard surfels $s$ whose support (i.e. closest) node ${support}(s)$ is in compression status.
(3) Discard surfels $s$ whose maximal distance to nearby depth surfels along the camera view direction is smaller than $\gamma_{inlier}$. All remaining candidates are appended to the latest geometry $S^t$. After new surfels are appended to the geometry, we collect surfels with shortest distances to the existing deformation node larger than $r_{sample}$. These surfels do not have a valid support node.  Thus, we perform a spatially uniform sampling from these surfels, and sampled nodes are appended to the deformation graph $G^t$.
Since new nodes are appended to the deformation graph, the historical maximum distance $d_h$ also needs to be updated. If $g_k$ belongs to the newly appended nodes, we initialize their $d_h$ by the following formula,
\begin{equation*}
   \forall g_j \in G^t, d_h^t(g_k, g_j) = max\{d_h^t(g_i, g_j) - d^t(g_i, g_k), d^t(g_j, g_k)\}.
\end{equation*}
wherein $g_i$ is the nearest neighbor node of $g_k$. According to the triangle inequality, $d_h^t(g_k, g_j)$ is a lower bound for the historical maximal distance between $k$ and $j$.

\textbf{Removal.} 
Surfels $s_{align}$ that are not registered are marked as removal candidates. There are three cases where a surfel $s$ should be removed: it is unstable for a long time (low confidence and not updated in a long time), overlapped with nearby surfels, or inconsistent with measurement $M$. 

A removal counter for each node is maintained to count how many of its supporting surfels have been removed. If this counter passes a threshold $\gamma_{remove}$, the removal of the corresponding node is triggered. Because it is very likely that there exists a topological change and tracking failure nearby this node, we perform a local re-initialization by removing this node and its supporting surfels.

After the update of geometry $S^t$ and deformation graph $G^t$ is finished, we update historical maximum distance $d_h^t$ between every two graph node with the follows:
\begin{equation*}
   \forall g_i,g_j \in \{G^t \cap G^{t-1}\}, d_h^t \leftarrow max(d_h^{t-1}, d^t(g_i,g_j)),
\end{equation*}
wherein $d^t$ is the Euclidean distance in $G^t$.

\subsection{Topological separation}
The final step of our system is the topological separation. This step separates the geometry $S$ and deformation graph $G$ into a set $\{(S_i,G_i)\}$, wherein $\{G_i\}$ are the connected components of $G$, which have been already separated based on the maximum historical distance $d_h$. These components are obtained by applying the  flood-fill algorithm on $G$.

\section{EXPERIMENTS}
\begin{figure}[ht]
 \center
  \includegraphics[width=0.48\textwidth]{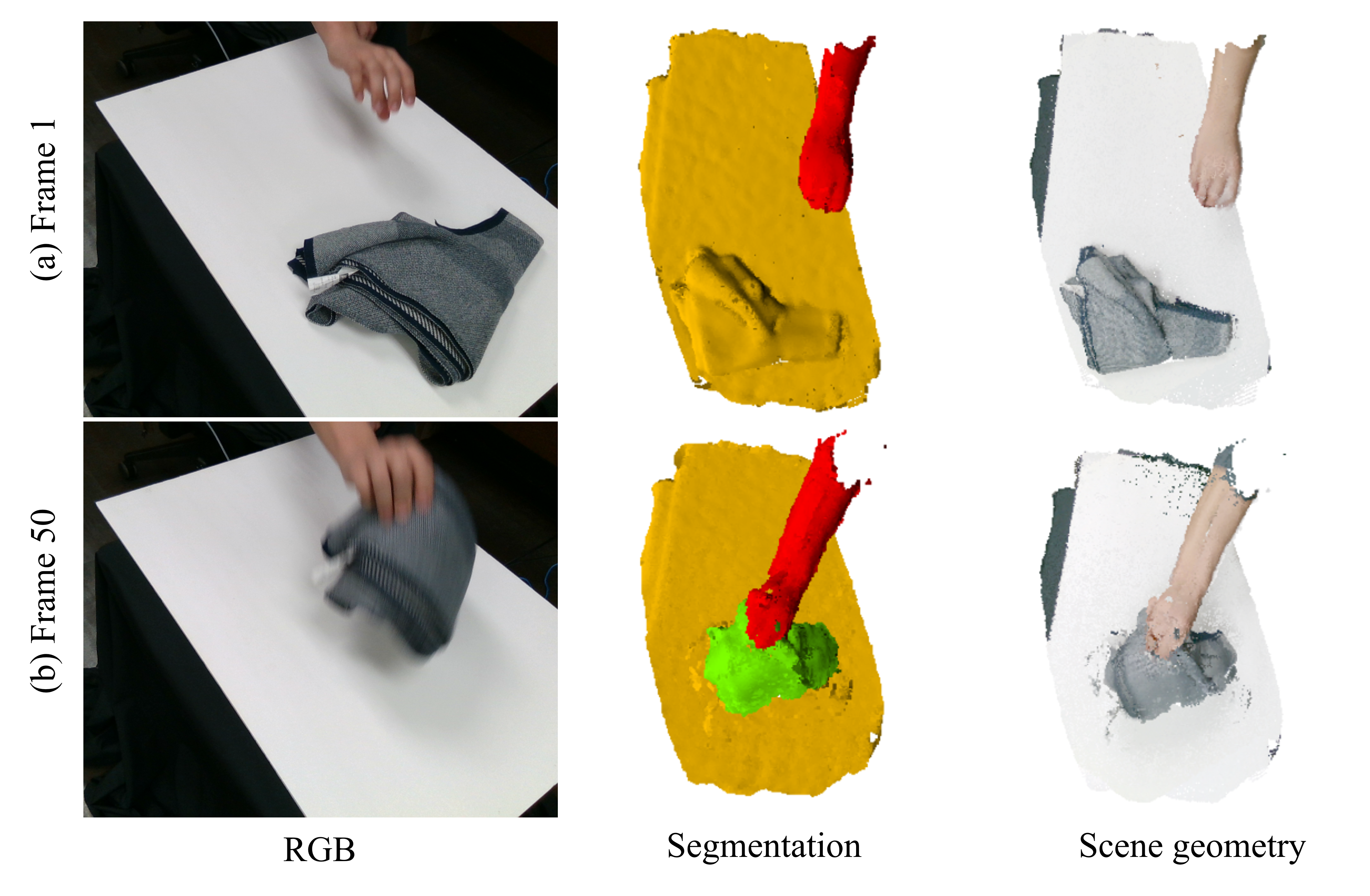}
  \caption{\small We pick a scarf from the table. The RGB images show how both our arm and the scarf are dramatically deformed. Despite that, our system successfully isolated each topology, tracked and reconstructed each object in the scene.}
  \vspace{-0.2cm}
  \label{soft_demonstration}
\end{figure}

\begin{figure}[ht]
 \center
  \includegraphics[width=0.48\textwidth]{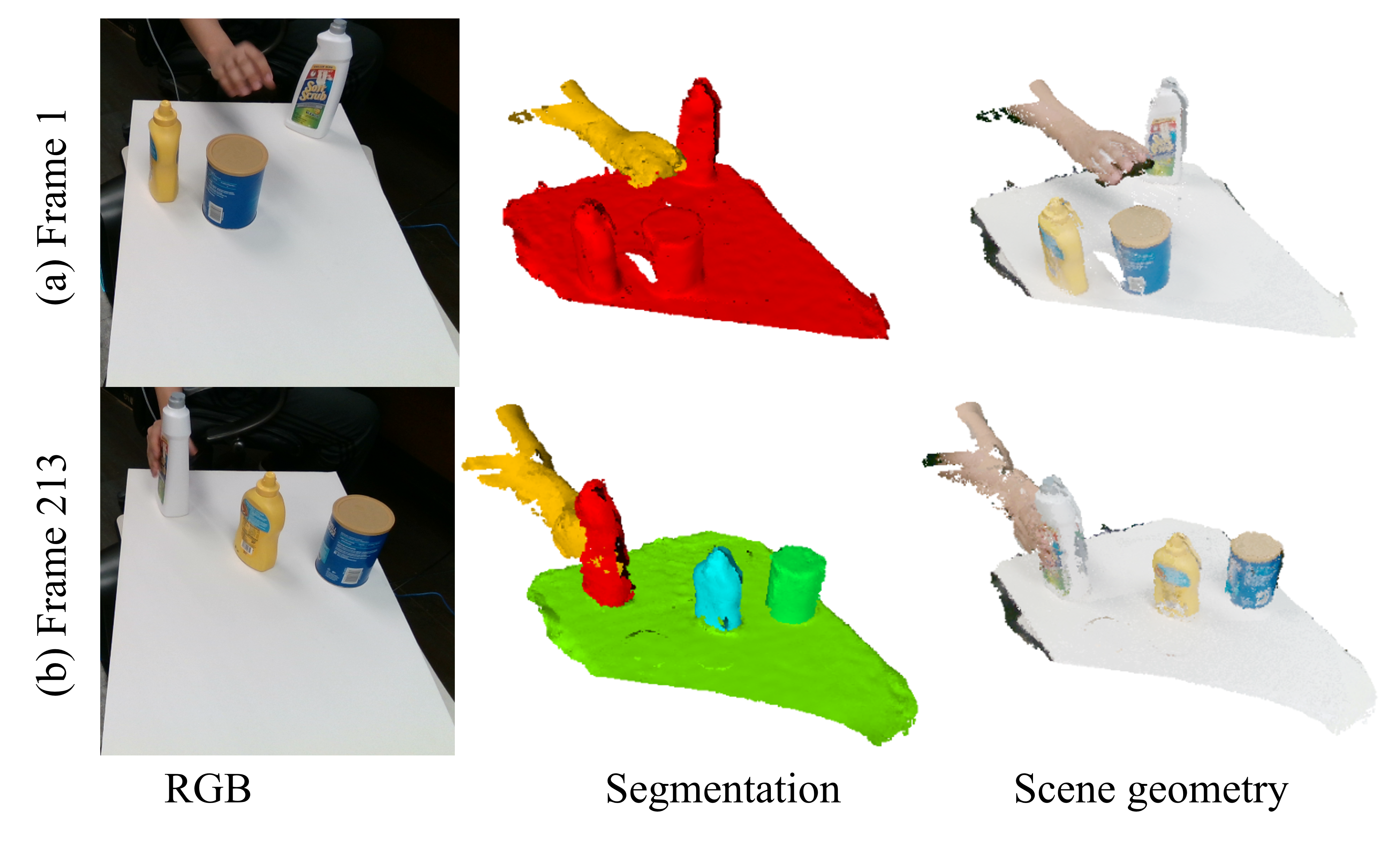}
  \caption{\small An object re-arrangement example. Three objects are moved to different positions in a 9-seconds video. This experiment demonstrates the capability of our system in dealing with multiple moving objects, and its potential use in robot re-arrangement tasks. }
  \vspace{-0.6cm}
  \label{multiple_demonstration}
\end{figure}

\begin{figure}[ht]
 \center
  \includegraphics[width=0.48\textwidth]{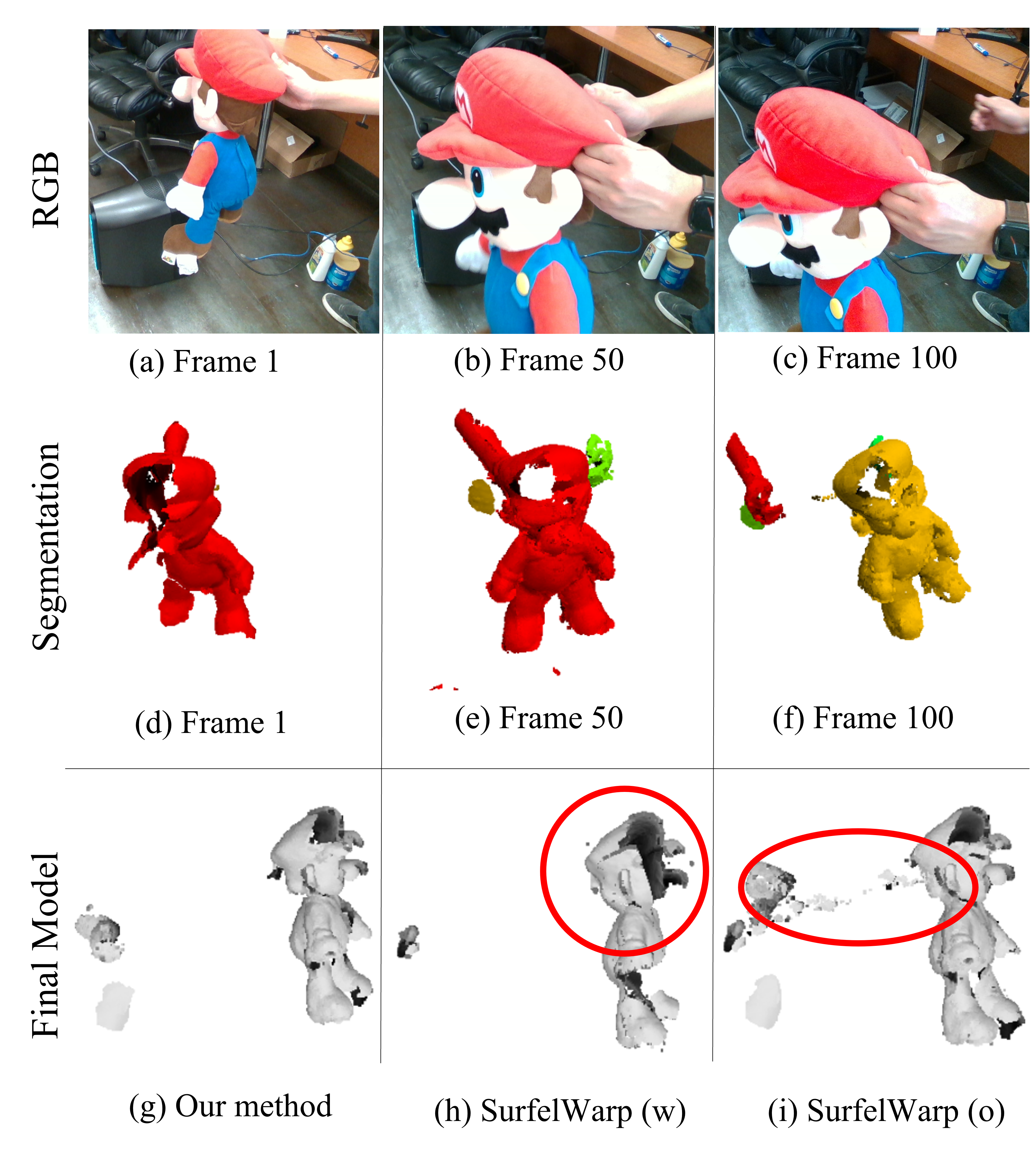}
  \caption{\small We move a Mario plush toy from one hand to another. Top row is the measurement from camera 0, and middle is object segmentation. Bottom row shows a result comparison among the our proposed method, SurfelWarp (w)~\cite{Gao2019} with global re-initialization, and SurfelWarp (o) without re-initialization. In the third row, we can observe that SurfelWarp (w) is losing too much information (e.g. a big hole in Mario's face). As for SurfelWarp (o), we observe that the hand and the toy are mistakenly connected. Our proposed method correctly separates the hand and the toy while maintaining most of the information.}
  \label{comparison}
\end{figure}

For non-rigid reconstruction problems, it is hard to provide quantitative results with real scenes~\cite{Newcombe,Gao2019}, because it is extremely challenging and expensive to obtain ground truth for deformable models. Thus, we conduct a set of qualitative experiments over a wide range of real-world scenarios that involve non-rigid object manipulation, which are closely related to robotic manipulation scenarios.
The experiments aim to assess the functionality of the proposed system and the validity of our novel topological change handling strategy.

Fig.~\ref{sequence_demonstrate} and Fig.~\ref{sequence_demonstrate_2} show the process of topological change handling.
For example, in Fig.~\ref{sequence_demonstrate}, four distinct topological changes are happening during this experiment; (1) the hand grasps the non-rigid plush toy, (2) the toy separates from the table, (3) the hand releases the toy, (4) and the toy rests on the table again. The Fig.~\ref{sequence_demonstrate_2} is even more complicated than Fig.~\ref{sequence_demonstrate}.
Our system detected and handled all of those topological changes correctly. 

Fig.~\ref{soft_demonstration} and Fig.~\ref{multiple_demonstration} demonstrate the stability of our proposed system under different challenging settings.
In Fig.~\ref{soft_demonstration}, our proposed framework successfully handles a scene with large deformation, picking up a scarf.
Fig.~\ref{multiple_demonstration} illustrates the capability of our system in processing multiple moving objects (5 objects in this scene, including the table), which demonstrates a potential usage in robot re-arrangement tasks.

Fig.~\ref{comparison} compares our reconstruction result with SurfelWarp~\cite{Gao2019} with/without global re-initialization, from which we can observe how global re-initialization loses too much information and no re-initialization cannot handle the topological change. This shows the superiority of the proposed local re-initialization strategy over the global re-initialization strategy. To the best of our knowledge, this is the first real-time non-rigid reconstruction system using local re-initialization as the topology handling strategy.

\textbf{Performance.}
The proposed system can run at $40$ fps. On average, the measurement fusion takes $4ms$, non-rigid alignment takes $10ms$, geometry and graph updates take $8ms$, and topology segmentation takes $3ms$. This performance is obtained by testing the system on a desktop machine with a GeForce RTX 3090 and an AMD-Ryzen 9 5900X, on the scene shown in Fig.~\ref{sequence_demonstrate}, which contains $13k$ surfels and $900$ deformation nodes. 

\textbf{Limitation and Discussion.} Although our system is able to solve the scene-level tracking and reconstruction problem without any prior, it still has many limitations. First, our current system relies on only depth measurement for registration and global registration is lacking here. If there is a fast and large deformation or motion in the scene, our system is not able to track it. How to combine global registration with our current system is a future direction of our work.
Our proposed pipeline relies heavily on measurements for surfel appending and removal. Although we are using a multi-view setting, we often lose the measurement of partially visible scenes. The reason is that our multi-view camera system is sparse (we only use 3 cameras in our experiment). For those areas missing measurement, we are unable to make any update or refinement towards their geometry. This is the reason why the edges of our reconstructed model are sometimes not smooth. 
Furthermore, if a local geometry is out of vision but deforms largely, likely, we have already lost track of it when it comes back to our view. This loop-closure problem is an ongoing topic in all works related to non-rigid scene reconstruction.

\section{CONCLUSION}
In this paper, we presented the first real-time solution for the STAR (Scene-level Tracking and Reconstruction) problem with no object prior, regardless of the rigidness or texture existence in the scene. Instead of segmenting the scene and reconstructing each object individually,   the operation order is reversed by reconstructing the non-rigid scene first and then performing topology-based segmentation. A novel surfel-based local re-initialization strategy is introduced to deal with frequent topological changes in the scene while maintaining most of the global geometry. The proposed method can be integrated seamlessly into several robotics applications, such as learning manipulation skills from visual demonstrations.

\bibliographystyle{IEEEtran}
\bibliography{Star} 
\end{document}